\documentclass[runningheads]{llncs}
\usepackage{graphicx}
\usepackage{comment}
\usepackage[ruled,vlined,linesnumbered]{algorithm2e}
\usepackage{eucal} 	 	
\usepackage{verbatim}      	
\usepackage{makeidx}       	
\usepackage{epsfig}         	
\usepackage{url}		

\begin{document}
\title{MAFAT: Memory-Aware Fusing and Tiling of Neural Networks for Accelerated Edge Inference\vspace{-5pt}}
\titlerunning{MAFAT: Memory-Aware Fusing and Tiling of Neural Networks}
%
\author{Jackson Farley \and Andreas Gerstlauer}
%
%
\institute{Electrical and Computer Engineering \\
The University of Texas at Austin, Austin, TX \\
\email{\{jackson\_farley,gerstl\}@utexas.edu}
\vspace{-10pt}}
\maketitle              
\begin{abstract}
A rising research challenge is running costly machine learning (ML) networks locally on resource-constrained edge devices. ML networks with large convolutional layers can easily exceed available memory, increasing latency due to excessive OS swapping. Previous memory reduction techniques such as pruning and quantization reduce model accuracy and often require retraining. Alternatively, distributed methods partition the convolutions into equivalent smaller sub-computations, but the implementations introduce communication costs and require a network of devices. Distributed partitioning approaches can, however, also be used to run in a reduced memory footprint on a single device by subdividing the network into smaller operations. 
In this paper, we extend prior work on distributed partitioning into a memory-aware execution on a single device. Our approach extends prior fusing strategies to allow for multiple groups of convolutional layers that are fused and tiled independently. This enables trading off overhead versus data reuse in order to specifically reduces memory footprint. 
We propose a memory usage predictor coupled with a search algorithm to provide optimized fusing and tiling configurations for an arbitrary set of convolutional layers. 
When applied to the YOLOv2 object detection network, results show that our approach can run in less than half the memory, and with a speedup of up to 2.78 under severe memory constraints. Additionally, our algorithm will return a configuration with a latency that is within 6\% of the best latency measured in a manual search. 

\keywords{Machine learning  \and Edge Computing}
\end{abstract}

\section{Introduction}
\index{Introduction@\emph{Introduction}}%



There has been a proliferation of complex machine learning (ML) problems in edge applications. Running ML applications 
on the edge can increase privacy, improve latency, reduce cloud communication, and require less energy~\cite{park2019edgegood}. 
However, most state-of-the-art ML networks have significant memory requirements that can exceed available memory on a resource-constrained edge device. Even with virtual memory enabled, exceeding memory bounds comes with severe latency penalties due to excessive swapping between memory and disk. As a result, it is a significant challenge to run networks locally on an edge device. 


Numerous commonly used neural networks contain a series of convolutional layers to process image data. Many convolutional layers, especially layers earlier in the network are feature-heavy, with a large amount of memory needed for inputs and outputs. Previous approaches to reduce memory footprints such as pruning ~\cite{Anwar15Pruning},~\cite{LiCNNPruning} and quantization~\cite{Gong14Quantization},~\cite{Lin15Quantization},~\cite{Laubeuf19Quantization} modify the network model, require re-training, and experience accuracy degradation. Meanwhile, distributed solutions such as~\cite{moDNN} and~\cite{deepthings} rely on partitioning convolutions into separate tasks and running them on separate devices, but they require additional communication and a network of devices. However, such approaches can also be used to reduce the memory footprint of a computation locally on a single device. 




In this paper, we extend the fused tile partitioning (FTP) approach outlined in~\cite{deepthings} to present a memory-aware fusing and tiling (MAFAT) strategy for the execution of large feature-dominated early stages of convolutional neural networks (CNNs) on a single resource-constrained edge device. 
The FTP approach from~\cite{deepthings} combines all layers into one large layer group and fuses them all together in order to reduce communication. By contrast, MAFAT creates two smaller layer groups and tiles and fuses them separately. The smaller fusings and different tilings resulting from more layer groups can reduce the maximum memory footprint of a process. 
We also develop a model to predict the maximum memory usage of a given MAFAT configuration. Finally, using this predictor, we propose a search algorithm that uses this predictor to return an optimized MAFAT configuration that fits within the provided memory requirement.

Results of applying our approach to a CNN used for object detection~\cite{yolov2} show that MAFAT configurations can provide a speedup of up to 2.78 over the original model in tighter memory constraints.  Furthermore, our search algorithm returns a configuration with a latency that is within 6 percent of the best measured latency for any configuration.

\section{Motivational Example}
\label{section:motivate}

\begin{figure}[tbp] 
\begin{center}
\psfig{file=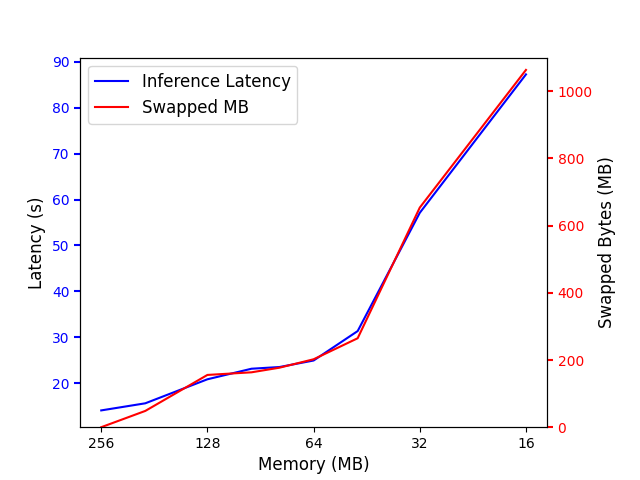, height=2in}
\vspace{-10pt}
\caption{The original YOLOv2 implementation for varying memory constraints.}
\label{fig:darknetMemory}
\end{center}
\end{figure}

Figure~\ref{fig:darknetMemory} depicts the latency and number of swapped bytes versus a decreasing memory constraint from running the first 16 layers YOLOv2~\cite{yolov2} on a Raspberry Pi3. The first 16 layers of the network are used because they are the most feature-heavy and present the greatest feature challenge to memory. Using MAFAT configurations on weight-heavy later layers will not have any added benefit and a single partition or other methods should be considered if these layers exceed memory requirements, which is out of the scope of this paper.
In addition to memory for weights and input and output features, Darknet allocates scratch space in order to do a layer calculation. This scratch space can go as high as 100MB for some layers. 
The largest combined memory is for layer 2. If that layer is loaded in its entirety, the processor needs at least 135 MB of memory for YOLOv2 to run cleanly. 

\index{commands!environments!figure}%

Figure~\ref{fig:darknetMemory} shows a significant increase in the latency of an inference at tighter memory constraints. The CNN exceeds memory constraints at over 192 MB. Once the program goes over memory, the OS must swap data between the memory and disk. This swapping process has a demonstrated adverse affect on latency. As the memory constraints continue to shrink, the inference latency increases dramatically, with a 16MB memory constraint over $6.5 \times$ slower than the original. This motivates a need for optimizations as presented in this paper to reduce the latency overhead due to swapping.




\section{Related Work}
The primary approaches to reduce memory on a single device are pruning and quantization. Pruning attempts to remove a portion of the model, such as weights in a filter, but this can result in asymmetric computations that can be difficult to implement~\cite{Anwar15Pruning}. Entire filters can be removed, too, such as in ~\cite{LiCNNPruning}. In both of these cases, pruning severely degrades accuracy and expensive retraining is required afterwards. Quantization of a CNN~\cite{Gong14Quantization},~\cite{Lin15Quantization} reduces the number of bits necessary to store weights. Similarly, retraining is often needed to get better accuracy~\cite{Laubeuf19Quantization}. Quantization also removes model information, i.e. it also degrades the accuracy of the model. By contrast, MAFAT is able to preserve model accuracy while decreasing the memory footprint. 
MAFAT is orthogonal to both pruning and quantization. Because the model is preserved, MAFAT can easily be applied to a pruned or quantized network. Combinations of MAFAT and quantization or pruning have the potential to shrink the memory footprint of convolutions significantly. Some prior work has considered memory balancing and swapping overhead when scheduling multiple DNNs on edge devices~\cite{Cox21Masa}. By contrast, MAFAT is aimed at reducing memory swapping overhead for each individual DNN. 

In addition to pruning and quantization, partitioning of models across multiple devices has been applied in distributed settings. For example, MoDNN~\cite{moDNN} uses a one-dimensional partitioning scheme where a map-reduce algorithm can execute many of the partitions in parallel. 
DeepThings~\cite{deepthings} uses Fused Tile Partitioning (FTP) to split layers into an even 2D grid and combines them via a fusing process in order for corresponding grid sections to be executed independently. Furthermore, DeepThings proposes data reuse and scheduling approaches such that adjacent partitions can use previously computed data where possible. However, all of these works are designed for computation among several devices. Because of this, communication is a primary consideration. 
Since MAFAT uses only a single device, alternative techniques such as partial fusing and re-tiling after a certain number of layers can result in more optimal memory usage.

\section{Memory-Aware Fusing and Tiling (MAFAT)}
\index{Memory-Aware Fusing and Tiling@\emph{Memory-Aware Fusing and Tiling}}%



This paper proposes memory-aware fusing and tiling (MAFAT)~\cite{myReport}, which builds on the fused tile partitioning (FTP) method from~\cite{deepthings}. 
FTP allows a set of convolutional layers to be split into multiple smaller sub-convolutions. Each sub-convolution consists of a tile of the original input and output feature maps, where the sub-convolutions combine and fuse corresponding tiles across all layers to execute as one unit.
Instead of fusing all layers to minimize communication, MAFAT separates layers into up to two layer groups to provide additional control over memory usage. 
For example, if the early layers take up significantly more data than the later ones, it may make sense to tile the earlier layers more heavily. In this case, there is less memory being used in the earlier layers, but there is no significant added overhead in later layers from unnecessary tiling. Additionally, for a smaller number of fused layers, the overlap incurred will be less. This means that there is less redundant computation, and the grid of earlier layers does not have large task size disparities. In a standard $3\times3$ fused tiling with data reuse, the middle task does not reuse any data. Because of this, it is much larger than the surrounding tiles and its memory usage is disproportionately larger. 


MAFAT currently takes any set of $n$ convolutional and maxpool layers. The layers are configured in a single layer group with all layers fused or two layer groups separated by a $cut < n$. This cut is the point at which the two layer groups are split. The first layer group will be from layer $0$ to layer $cut-1$, and the second rom layer $cut$ to layer $n$. In this way, each layer is part of one of the two layer groups. There is some additional overhead for storage of additional parameters be stored, and the cut layer must be merged in memory and re-tiled.

Potential cuts are determined in a memory-aware fashion. Collecting all the tiled data into a single input tensor and re-tiling can be memory intensive. To make this as efficient as possible, the cuts were chosen to be directly after maxpool layers. After these layers, the tensors are significantly smaller, as they have effectively just been down-sampled. In the YOLOv2 example, these potential cuts are at layers 2, 4, 8, and 12.
For the two layer groups, the tiling for each group is independent of the other. This means that the first layer group could be tiled at $5\times5$ while the second could be tiled at $2\times2$. The potential tilings were all even on height and width, and were $1\times1$, $2\times2$, $3\times3$, $4\times4$, and $5\times5$.




\subsection{Predicting Maximum Memory Usage}

We also developed a predictor of the maximum memory usage of a given MAFAT configuration based on the maximum memory usage of the largest tile in each layer group. Layer groups generally exceed memory towards the beginning and middle of their execution. 
It was found that the factors that best predicted maximum memory usage were the largest combination of: (1) scratch space of tile $t$, (2) input to tile $t$, (3) output of tile $t$, and (4) output of previous layer to tile $t$.
While other parameters such as the size of data reuse and size of tasks waiting in the processing queue were considered, these were found to negatively affect the ability of the predictor to accurately predict memory usage. Additionally, weights for all layers in the fusing are assumed to be in memory constantly, as well as a significant amount of additional overhead devoted to network parameters, system variables, and other data. A constant $bias$ term of 31 MB was empirically determined to account for these. This bias depends on the operating system, network and hardware platform.

\begin{algorithm}[t]
\upshape{predictLayerGroup}$(N,M,\mathbf{W},\mathbf{H},\mathbf{F},\mathbf{S},top,bottom)$

$max \leftarrow 0$\;

\For{$i \in 0..N$}{
    \For{$j \in 0..M$}{
        $l \leftarrow bottom$\;
        \While{$l \le top$}{
            $w_{in}, h_{in}, w_{out},h_{out},c_{in},c_{out} \leftarrow$ \upshape{Grid}$(l,N,M,W_l,H_l,i,j)$\;
            $scratch \leftarrow w_{out} \times h_{out} \times c_{in} \times (F_l)^2 / S_l$\;
            $input \leftarrow w_{in} \times h_{in} \times c_{in}$\;
            $output \leftarrow  w_{out} \times h_{out} \times c_{out}$\;
            $mem \leftarrow scratch + output + (input\times2)$\;
            \If{$mem > Max$}{
                $Max \leftarrow mem$\;
            }
            $l \leftarrow l - 1$\;
        }
    }
}
\upshape{return}~$Max + bias$\;

 \caption{Memory predictor for a single layer group\label{algo:ftppredict}}
\end{algorithm}

A memory prediction is obtained as the maximum over predicted memory usage for all tiles in all layer groups in a given cut configuration. 
The memory predictor for one layer group is shown in Algorithm~\ref{algo:ftppredict}. This algorithm predicts the maximum memory usage of a given layer group and tiling strategy. The inputs to Algorithm~\ref{algo:ftppredict} are the parameters of a layer group spanning from layer $top$ to layer $bottom$ with an $N\times M$ tiling strategy, as well as a network configuration $\mathbf{W},\mathbf{H},\mathbf{F},\mathbf{S}$ with each layer $l$ having width $W_l$ and height $H_l$, filters of size $F_l$ and a stride of $S_l$. The stride is how much the filter moves each computation.
The \textit{Grid} function in Algorithm~\ref{algo:ftppredict} calculates and returns
the dimension of a tile in layer $l$ including additional overlap following the traversal function in~\cite{deepthings}.



\begin{figure}[tbp]
\vspace{-10pt}
\begin{center}
\psfig{file=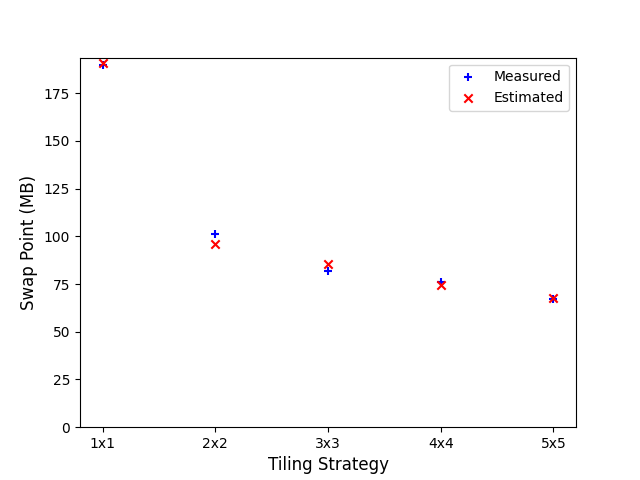,width=0.49\columnwidth}
\psfig{file=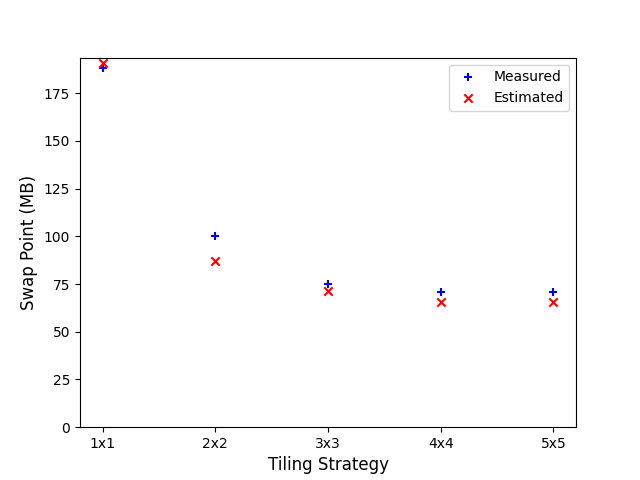,width=0.49\columnwidth}
\vspace{-10pt}
\caption{Memory usage prediction for fully fused 16 layers (left) and for 8 fused layers with a 2x2 fused tiling on layers 9-16 (right).}
\label{fig:swapPredictor}
\end{center}
\end{figure}

Figure~\ref{fig:swapPredictor} depicts the predicted memory limit and the measured limit for a single layer group and for the MAFAT configurations with a cut at layer 8 and a $2\times2$ bottom tiling strategy.
The measured limit was determined using the setup in Section~\ref{section:results} by decreasing the memory constraint in 1 Megabyte increments until swaps were observed. The predictor performs well in both cases.


\subsection{Configuration Algorithm}

To determine the ideal MAFAT configuration, Algorithm~\ref{algo:getConfig} performs a greedy search over a subset of the configuration space to find the configuration with the fewest tiles. Its goal is to return a near-optimal configuration of the network such that the end latency will be as small as possible. The inputs to the algorithm are the layer parameters and the memory limit. The relevant layer parameters are width ($W_l$), height ($H_l$), filter size ($F_l$), stride ($S_l$), and the total number of layers to be fused ($n$). The vector of potential cuts ($Cuts$) here is specific to YOLOv2, due to the location of the maxpool layers. A further restriction of the search space is based on a manual configuration. Specifically, no latency advantage was found for cuts at layer 4, and when there were cuts made, the best performing second layer group tiling was $2\times 2$. The tiling strategies are also currently limited to even squares. 

Algorithm~\ref{algo:getConfig} returns the number of tiles for the first layer group $LG_1$, the cut $cut$, and the tiling for the second layer group $LG_2$. It performs the modified search starting at the highest memory value, and slowly creates more even configurations that require more overhead, but fit in smaller memory footprints. If a configuration is found that fits in the memory limit, there is no unexplored configuration in the search space that will produce a higher memory prediction. Therefore, the latency returned should be the lowest. If virtual memory is enabled, this algorithm assumes that any additional swaps from the operating system will be slower than picking a better configuration. If no configuration can be found, then the algorithm returns the most even configuration: $5\times5$ into $2\times2$ with a cut at layer 8. 

\begin{algorithm}[tb]
\upshape{getConfig}$(\mathbf{W},\mathbf{H},\mathbf{F},\mathbf{S},n,MemoryLimit)$\\
 $Cuts \leftarrow \{16,12,8\}$\;
 $Tiles \leftarrow \{1, 2, 3, 4, 5\}$\;
 $LG_2 \leftarrow 4$\;
 $N_2 \leftarrow LG_2;~ M_2 \leftarrow LG_2$\;
 $l \leftarrow getMaxLayer(NetworkParams)$\;

 \For{$cut \in Cuts$}{
    \For{$tile \in Tiles$}{
        $LG_1 \leftarrow tile$\;
        $N_1 \leftarrow LG_1;~ M_1 \leftarrow LG_1$\;
        \If{\upshape{predictMem}$(N_1,M_1,N_2,M_2,\mathbf{W},\mathbf{H},\mathbf{F},\mathbf{S},cut,n) < MemoryLimit$} {
            \textbf{\upshape return} $LG_1$, $LG_2$, $cut$\;
        }
    }
 }

 \textbf{\upshape{return}}~$LG_1,LG_2,cut$

\caption{Configuration search algorithm\label{algo:getConfig}}
\end{algorithm}

\section{Experimental Results}
\index{Experimental Results@\emph{Experimental Results}}%
\label{section:results}

We applied MAFAT to the YOLOv2 object detection network. The measurements were all carried out on a Raspberry Pi3 running Raspian. The Raspberry Pi was equipped with a quad-core 1.2GHz ARM Cortex-A53 processor and a total memory size of 1 GB. During the measurements, we restricted the Raspberry Pi to a single core and a variable amount of memory from 16MB up to 256MB. 


A separate measurement thread was created to measure system swaps in and out of memory each second. This gives information about likely places for a bottleneck. This was achieved using the \verb|vmstat| command. Due to the \verb|vmstat| only working at a full system level, it was crucial to keep the test environment free of as many other running processes as possible. Despite this, there is some noise in the swap measurement.
 
To measure memory usage of just the process, an additional thread was used that polled the process using the \verb|ps| command. This way we could filter out other processes without as much added system noise. This was useful in seeing more accurately where the swapping would line up with the program. 
 
Both of these threads however added some additional memory usage and could potentially increase swaps or create conflicts with the process. Therefore, when the latency for the process was calculated, internal measurements were used via the \verb|chrono.h| library in C++ for accurate, wall clock times at a millisecond granularity. This also allowed for precise measurements at the beginning and end of an inference. In this paper, the latency was measured before the input image was loaded and after the first 16 layers had executed. 

To mimic a smaller edge device with minimal effort, this paper used control groups. Specifically, the \verb|cpuset| and \verb|memory| control groups were used to restrict the experiment to a single core and a smaller amount of memory, respectively. 
This allowed for finer adjustments of memory constraints without the need for rebooting. 
For predictability and reproducibility, as few active processes as possible were running during final latency measurements. 

\subsection{Manual Exploration}
To develop the algorithm, and to better understand the configuration performance, we first performed a manual search of different possible configurations. In the following, a MAFAT configuration with a top layer group tiling of $N_1\times M_1$, a cut a layer $c$, and a bottom layer group tiling of $N_2 \times M_2$ is written as $N_1\times M_1/c/N_2\times M_2$
Using prior knowledge, the search space of possible cuts was restricted. As mentioned before, intermediate data is reduced the most by cutting the network into two layer groups at layers 4, 8, and 12, or no cut at all. In each case, all layers up to and after the cut were fused together. Additionally, the final layers were split into either $2\times2$ or $3\times3$ tiles for reducing maximum memory while still allowing for faster processing times. The tilings for the top layer group were swept from $1\times1$ to $5\times5$. 

\begin{figure}[tbp]
\begin{center}
\psfig{file=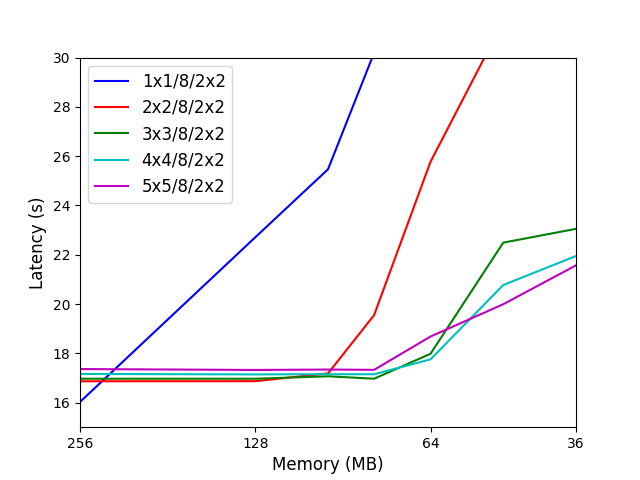,width=0.49\columnwidth}
\hfill
\psfig{file=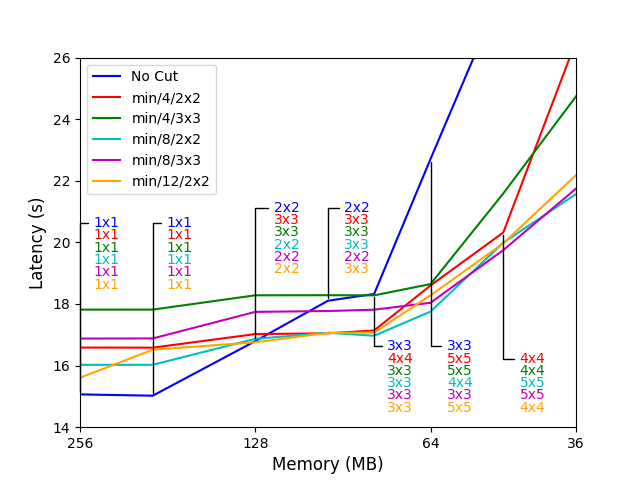,width=0.49\columnwidth}
\vspace{-10pt}
\caption{Latency for different tilings cut at layer 8 (left) and for different cuts (right).}
\label{fig:allConfigs}
\end{center}
\end{figure}

Figure~\ref{fig:allConfigs} shows the effect of top and bottom layer group tiling strategies on measured latency across a shrinking memory limit. 
In the graph on the left, 
each line represents the tiling of the top layer group, which is then cut at layer 8 and fed into a $2\times2$ bottom layer group. Results demonstrate the superiority of finer tilings in smaller memory footprints, but also the additional overhead they generate when more memory is available. For high memory values in excess of 200 MB, the 1x1 tiling is best. On the other hand, using a $4\times4$ or $5\times5$ tiling scheme yields much better results for lower memory values. 
 
The graph on the right shows the effect of cut placement and bottom layer group tiling strategy. The top tiling for this line is the tiling strategy (from $1\times 1$ to $5\times 5$) that yields the smallest latency for the given cut and bottom tiling. The best top tiling for each configuration is also annotated onto the graph at each memory point. For example, the $min/8/3\times3$ line represents a cut at layer 8 with the best top tiling and a $3\times 3$ bottom tiling. It can therefore be viewed as the optimized top tiling for a given cut and bottom tiling. As seen in the graph, middle cuts at layer 8 have the fastest latency at tighter memory restrictions. It is also clear that the absence of a cut becomes costly at tighter restrictions due to additional layer overlapping. This figure also reinforces top tiling results to show that finer tilings perform better at tighter memory restrictions.



\begin{figure}[tb]
\begin{center}
\ \psfig{file=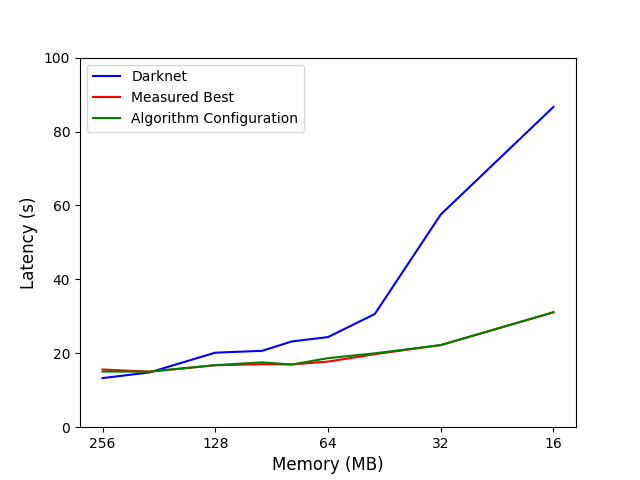,height=2in}
\vspace{-10pt}
\caption{Darknet latency compared to algorithm and minimum latency measured.}
\label{fig:darknetVsBest}
\end{center}
\end{figure}

Figure~\ref{fig:darknetVsBest} compares the best measured latency obtained by the MAFAT manual exploration and search algorithm to the original latencies measured from the standard Darknet implementation across decreasing memory limits.
It is clear that MAFAT outperforms Darknet and reduces the latency and swaps. 
Interestingly, the minimum configuration for the algorithm, $5\times5/8/2\times2$, is predicted to have a maximum memory usage of 66 MB. Currently, therefore, there is not a MAFAT configuration that does not run in less than a 66 MB footprint without swapping. However, as memory restrictions get even tighter, the latency increases at a much slower rate than Darknet. This shows that the MAFAT configuration also performs much better under swapping due to more even memory usage across the execution of the network.



\index{commands!environments!figure}%

\begin{table}[bp]
\caption{Comparison of configurations and latencies.}
\label{tab:configCompare}
\centering
\vspace{-5pt}
\begin{tabular}{|r|c|c|c|c|}
\hline
& \multicolumn{2}{c|}{Best Measured} & \multicolumn{2}{c|}{Algorithm} \\ \cline{2-5}
MB & Configuration  & Latency (ms)  & Configuration & Latency (ms)  \\ \hline
256 & 1x1/NoCut  & 15065 & 1x1/NoCut  & 15065 \\
192 & 1x1/NoCut  & 15023 & 1x1/NoCut  & 15023 \\
128 & 2x2/12/2x2 & 16757 & 2x2/NoCut  & 16795 \\
96  & 3x3/4/2x2  & 17048 & 2x2/12/2x2 & 17543 \\
80  & 3x3/8/2x2  & 16968 & 3x3/8/2x2  & 16968 \\
64  & 4x4/8/2x2  & 17753 & 5x5/8/2x2  & 18679 \\
48  & 5x5/8/3x3  & 19749 & 5x5/8/2x2  & 19991 \\
32  & 5x5/8/2x2  & 22215 & 5x5/8/2x2  & 22215 \\
16  & 5x5/8/2x2  & 31095 & 5x5/8/2x2  & 31095 \\
\hline
\end{tabular}
\end{table}

\subsection{Algorithm Performance}

Figure~\ref{fig:darknetVsBest} also plots the measured performance for the configurations produced by our optimization algorithm. The differences between the algorithm and the best measured are shown to be minimal. The algorithm's specific configuration compared to the best measured can be found in Table~\ref{tab:configCompare}. To evaluate algorithm performance, the outputs of the algorithm were calculated for the memory values in the table. This allowed for easy comparison with the existing measured data. Notably, the latency values are quite similar and are all within 6 percent of the best measured from manual exploration. Given how the algorithm relies on prior knowledge and some of the data already recorded, this level of performance is not surprising. However, the intuition behind the algorithm and the basic results should help apply it in other domains. 


\index{commands!environments!table}%
\section{Summary and Conclusions}
\index{Summary and Conclusions@\emph{Summary and Conclusions}}%

This paper presents memory-aware fusing and tiling (MAFAT), an expansion of existing fusing and tiling strategies in order to make feature-heavy convolutional neural network layers feasible on smaller edge devices. Originally, edge devices would have increasing latency measurements due to swapping data between the memory and disk. Many edge devices cannot spare 200 MB to run early convolutional layers, so we break up each layer into sub-convolutions that can then be grouped together and executed in a much smaller memory footprint. This paper shows that certain configurations of tiling can offer a respectable 1.37 speedup compared to the naive approach at 64 MB and up to a 2.78 speedup with only 16 MB available. 
Additionally, the intuition and structure behind the memory usage of the process is explored, and a simple algorithm is proposed to predict the maximum memory usage of a MAFAT configuration. Given this, an appropriate configuration can be returned for a user to use that is within 6 percent of the best measured latency from a manual exploration.
The code used to take these measurements can be found at~\cite{myCode}. This research area can be further improved by use variable tiling, where each end tile is not the same size to allow for reduced task size variation and thus smaller footprints. We also want to generalize this algorithm to other tiling strategies and other CNNs. Currently, the end user must pre-determine what cuts make sense. 

%
%
%
\bibliographystyle{splncs04}
\bibliography{diss}
%
%
%
%
%

\end{document}